# Detecting Reddit Users with Depression Using a Hybrid Neural Network SBERT-CNN

Ziyi Chen
*Department of Biostatistics*
*University of Washington*
Seattle, WA, USA
zychen09@uw.edu

Ren Yang
*Departemnt of Artificial*
*Intelligence and Informatics*
*Mayo Clinic*
Rochester, MN, USA
Ren.Yang@mayo.edu

Sunyang Fu
*Departemnt of Artificial*
*Intelligence and Informatics*
*Mayo Clinic*
Rochester, MN, USA
Fu.Sunyang@mayo.edu

Nansu Zong
*Departemnt of Artificial*
*Intelligence and Informatics*
*Mayo Clinic*
Rochester, MN, USA
Zong.Nansu@mayo.edu

Hongfang Liu
*Departemnt of Artificial*
*Intelligence and Informatics*
*Mayo Clinic*
Rochester, MN, USA
Liu.Hongfang@mayo.edu

Ming Huang
*Departemnt of Artificial*
*Intelligence and Informatics*
*Mayo Clinic*
Rochester, MN, USA
huang.ming@mayo.edu

*Abstract*—Depression is a widespread mental health issue, affecting an estimated 3.8% of the global population. It is also one of the main contributors to disability worldwide. Recently it is becoming popular for individuals to use social media platforms (e.g., Reddit) to express their difficulties and health issues (e.g., depression) and seek support from other users in online communities. It opens great opportunities to automatically identify social media users with depression by parsing millions of posts for potential interventions. Deep learning methods have begun to dominate in the field of machine learning and natural language processing (NLP) because of their ease of use, efficient processing, and state-of-the-art results on many NLP tasks. In this work, we propose a hybrid deep learning model (SBERT-CNN) which combines a pretrained sentence BERT (SBERT) and convolutional neural network (CNN) to detect individuals with depression with their Reddit posts. The sentence BERT is used to learn the meaningful representation of semantic information in each post. CNN enables the further transformation of those embeddings and the temporal identification of behavioral patterns of users. We trained and evaluated the model performance to identify Reddit users with depression by utilizing the Self-reported Mental Health Diagnoses (SMHD) data. The SBERT-CNN model achieved an accuracy of 0.86 and an F1 score of 0.86 and outperformed the state-of-the-art documented result (F1 score of 0.79) by other machine learning models in the literature. The results show the feasibility of the hybrid SBERT-CNN model to identify individuals with depression. Although the hybrid model is validated to detect depression with Reddit posts, it can be easily tuned and applied to other text classification tasks and different clinical applications.

## I. Introduction

Depression is one of the most common mental illnesses and the influence of depression on society is a serious concern for public health communities. American Psychiatric Association (APA) approximates that 6.9 percent of the population in the United States is currently affected [1], costing the United States $233 billion in 2016 [2]. The World Health Organization (WHO) estimates that depression affects around 322 million people, at a cost of approximately 2.5 trillion dollars worldwide in 2017 [3]. WHO reported that the incidence of depression increased an extra 25% in the first year of the COVID-19 pandemic, mostly due to the aspects of social lockdown and isolation and work reduction [4]. Depression characterized by a sad, empty, or irritated mood with somatic and cognitive changes can worsen the individuals' capacity to function, limiting performance in everyday duties and social life [5]. Depression is a leading cause of disability around the world [2] and depression can lead to suicide, one of the leading causes of death [6]. Moreover, it is estimated that around two-thirds of depression cases are undiagnosed. The undiagnosed and thus untreated depression have a negative impact on quality of life and workplace productivity [7]. Prevention and intervention strategies such as screening and early detection are required to reduce the impact of depression.

Recently it is becoming popular for individuals to use social media platforms (e.g., Reddit) to express their life difficulties and health issues (e.g., depression) and seek support from other users in online communities [8, 9]. The social media platforms allow the public to anonymously disclose personal information and discuss life experiences and problems (e.g., depression) that may be stigmatizing, with less fear of offline harm or consequences [10, 11]. Recently research studies also confirmed that social media is effective platforms for self-disclosure and social support seeking for mental health difficulties (e.g., depression) [10, 12]. This makes social media platforms including Reddit important resources for studying user narratives related to depression [13, 14].

Among these social media platforms, Reddit ranks the 5th most visited website in the United States, having 52 million users per day and more than 430 million active users per month [15, 16]. Reddit consists of network of communities (subreddits) with focuses on specific topics such as depression, which enables users to interact and share with other individuals with similar backgrounds, views, and life experiences [17]. As of 2021, the platform had 2.8 million subreddits [18]. Such a large number of subreddits with posts in the Reddit provides us a valuable resource to identify individuals with depression for intervention [13]. However, it would be very challenging for mental health providers to review all the posts and replies to identify the individuals who suffered from depression due to the large volume of Reddit posts. Thus, it opens great opportunities to develop artificial intelligence (AI) and natural language processing methods (NLP) methods for automatically parsing millions of social media posts and identifying the users with depression for intervention.

There have been many existing works on detecting depression from social media posts with NLP algorithms, machine learning methods, deep learning techniques [13, 19, 20]. In the earlier works, traditional machine learning methods such as Logistic Regression, Random Forest, and Support Vector Machine (SVM) were used for identifying depression with features (e.g., n-grams, sentiment, and social characteristics) from extensive feature engineering. [21] Recent success in the use of deep learning for NLP tasks have motivated the identification of depression from social media posts by these deep learning techniques including Convolutional Neural Network (CNN) [22-24], Recurrent Neural Networks (RNN) [23, 25], and Transformers [26, 27].

In this paper, we propose a hybrid neural network model (SBERT-CNN) that concatenates a pre-trained sentence BERT (SBERT) model [28] and a CNN model for identifying Reddit users with depression. The SBERT model was applied to learn semantic representation of each post. CNN was used for the transformation of those embeddings from the pretrained models and the identification of behavioral patterns of users from transformed embeddings. The SBERT-CNN model was then trained and evaluated by using the Self-reported Mental Health Diagnoses (SMHD) data, a large dataset of social media posts from users with mental health conditions (e.g., depression) along with matched control users [24]. Our results were compared with the performance of other machine learning models trained with the same SMHD dataset in the literature. The comparison revealed very encouraging results that the hybrid SBERT-CNN model outperforms existing solutions for automatic identification of depression in the same dataset. Our findings suggest the feasibility of the hybrid SBERT-CNN model for detecting depression from social media posts. The proposed model can also be easily tuned and applied to other text classification tasks and to different clinical applications such as anxiety identification.

The remainder of this work is divided into five sections. Section II introduces the relevant machine learning methods trained and evaluated with the SMHD dataset. Section III describes the experimental data that were used and the structure of SBERT-CNN in details. Section IV shows the model performance and experiment results. Section V discussed our model with the existing machine learning methods and future work. Section VI concludes this work.

## II. RELATED WORK

Several studies have contributed to the classification of mental disorders, particularly depression, with the data from social media. In this section, we will summarize the existing machine learning methods developed for depression identification and validated with the same SMHD dataset.

An important factor for developing robust mental health classifiers is the creation of a large, annotated corpus as training and testing machine learning models. In 2017, Yates A et al. developed a automatic labeling technique for annotating Reddit users with depression and created a large size general forum dataset comprising users with self-reported depression diagnoses (RSDD) paired with control users [22]. The RSDD excludes all obvious Reddit posts on depression. They proposed a CNN architecture for depression detection which got F1 score of 0.51.

Later in 2018, Arman Cohan et al. applied the automatic labelling technique to other mental health disorders and created a new large Self-reported Mental Health Diagnoses (SMHD) dataset from Reddit posts [24]. The SMHD dataset contains public Reddit posts from users labeled with mental health conditions together with matched control. They used traditional machine learning techniques (e.g., Logistic Regression, SVM, and XGBoost), CNN, and FastText [29] to develop a binary classifier to identify individuals with depression. The best performance was achieved by supervised FastText with a F1 score of 0.54.

In 2019, Strube et al. employed the Hierarchical Attention Network (HAN) for depression identification with SMHD dataset and achieved a F1 score of 0.68 [30]. In 2021, Dinu et al. used three pretrained transformers - BERT, XLNET, and RoBERTa to develop a binary classifier in the post level with SMHD dataset and obtained a F1 score of 0.68, 0.70, and 0.68, respectively [26]. In 2022, Vanessa et al. deployed word embedding techniques including GloVe and Word2Vec to generated domain-specific embeddings of Reddit posts in the SMHD dataset [23]. These embeddings are imported into CNN and LSTM models to detect depression from Reddit posts, obtaining a F1 score of 0.79 and 0.77, respectively.

The good performance of pretrained models and CNN with embeddings inspired the proposal and exploration of a hybrid neural network model that concatenated sentence BERT and CNN for identifying Reddit users with depression. The SBERT model allows the language learning of semantic representation of each post. CNN enables the transformation of those embeddings learned by the pretrained models and the identification of behavioral patterns of users from semantic embeddings.

## III. METHODS

### A. Data

The Self-reported Mental Health Diagnoses (SMHD) dataset includes public Reddit posts from individuals who reported having been diagnosed with one or more mental health conditions including depression, as well as matched control users between January 2006 and December 2017. With a macro-averaged precision of 95.8% and a minimum precision of 90% for anxiety, this dataset achieves a high level of accuracy [24]. We utilized the SMHD dataset to derive the datasets on depression and the respective control groups to train and evaluate the SBERT-CNN model to identify Reddit users with depression. The original SMHD dataset randomly divides individuals and their posts into three subsets - training, validation, and testing sets, for model training, tuning, and testing, respectively. We kept the original instance separation and extracted depression-related data along with the matched control groups randomly sampled. Table I shows detailed information about the three depression datasets that we derived.

The training set comprises 216,022 posts, with an average of 45 tokens per post and a 95th percentile of 161 tokens per post, contributed by 1,316 users with depression. In comparison, the control group includes 393,449 posts, with an average of 26 tokens per post and a 95th percentile of 89 tokens per post written by an equal number of users. Similarly, the validation set includes 290,858 posts, with an average of 46 tokens per post and a 95th percentile of 162 tokens per post posted by 1,308 users with depression. On the other hand, the control group has 393,930 posts, with an average of 27 tokens per post and a 95th percentile of 91 tokens per post written by an equal number of users. Finally, the testing set contains 209,188 posts, with an average of 44 tokens per post and a 95th percentile of 160 tokens per post authored by 1,316 users with depression. In comparison, the control group includes 390,385 posts, with an average of 27 tokens per post and a 95th percentile of 91 tokens per post written by an equal number of users.

TABLE I. DEPRESSION DATASET

| Datasets | Labels | Total Users | Total Posts | Average Tokens | 95% tokens |
|---|---|---|---|---|---|
| Train | Depression | 1,316 | 216,022 | 45 | 161 |
| | Control | 1,316 | 393,449 | 26 | 89 |
| Valid | Depression | 1,308 | 290,858 | 46 | 162 |
| | Control | 1,308 | 393,930 | 27 | 91 |
| Test | Depression | 1,316 | 209,188 | 44 | 160 |
| | Control | 1,316 | 390,385 | 27 | 91 |

### B. Data Preprocessing

To prepare the depression dataset for deep learning models, we cleaned and processed the Reddit posts in the dataset before transforming them into numerical embeddings. This data preprocessing enables the more efficient and accurate transformation of text into computational vectors, ultimately enhancing the overall performance of deep learning models. The first step is to remove noise from the original text to improve the purity of the text. We eliminated any unreadable noise characters such as URLs, emojis, and other non-ASCII characters in the Reddit posts. We converted all the characters in the Reddit posts to lowercase to prevent case sensitivity. We expanded each contraction into an original phrase or multiple terms using the contractions package to reduce their ambiguity. After that, the cleaned text was tokenized into individual words using the NLTK package [31], and the first 512 words were kept and combined as input for pretrained models. SBERT is used as pre-trained language model for converting each post into a numerical vector, which allows a maximum of 512 tokens in an input.

### C. SBERT-CNN

Figure 1 illustrates the architecture of SBERT-CNN, which concatenates sentence BERT (SBERT) and CNN after text processing. We will discuss SBERT and CNN in more details in the following sections.

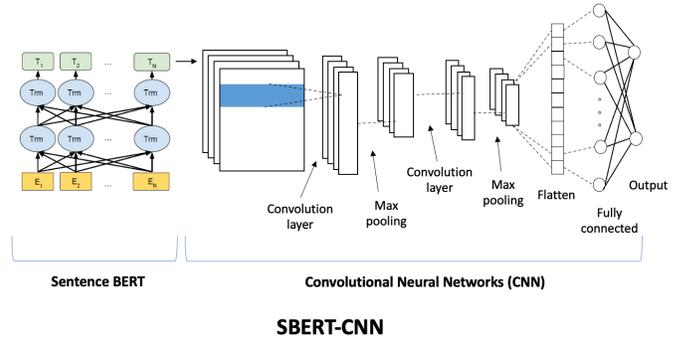

Fig. 1. Architecture of SBERT-CNN

### D. Sentence BERT (SBERT)

Recent advances in deep learning methods (e.g., transformer architectures and large-scale pre-trained language models) push the state-of-the-art for multiple NLP tasks [32]. The pre-trained language models enable the effective learning of language representation from a massive collection of documents and the generation of meaningful semantic embeddings for downstream NLP tasks. With semantic embedding that encodes words, sentences and their semantic content into numerical vectors, the machine can then learn and understand the context, intent, and other subtleties in the overall text with the assistance of this technique.

We used the pre-trained sentence BERT to transform each post into a numerical vector of semantic embedding. Sentence BERT was developed in 2018 and quickly gained the edge in semantic embeddings of sentences [28]. It is a variation of the conventional pre-trained BERT network that uses Siamese and triplet networks to generate sentence embeddings. A pre-trained mini model 'all-MiniLM-L6-v2' was used and other larger models such as 'all-mpnet-base-v2' will be explored in future work. Each post of users was converted into an embedding using the pre-trained model SBERT, resulting in an embedding matrix per user as an input of CNN.

### E. Convolutional Neural Network (CNN)

Convolutional Neural Network (CNN) is a type of artificial neural network inspired by the biological process of neurons in the visual cortex and features a convolution operation in place of general matrix multiplication [33-35]. CNN allows effective feature learning from raw data through a multi-layer data representation architecture, in which the first layers extract the low-level features while the last layers extract the high-level

Identify applicable funding agency here. If none, delete this text box.

features [36]. CNN has shown promising results for computer vision and NLP including text classification [37-39].

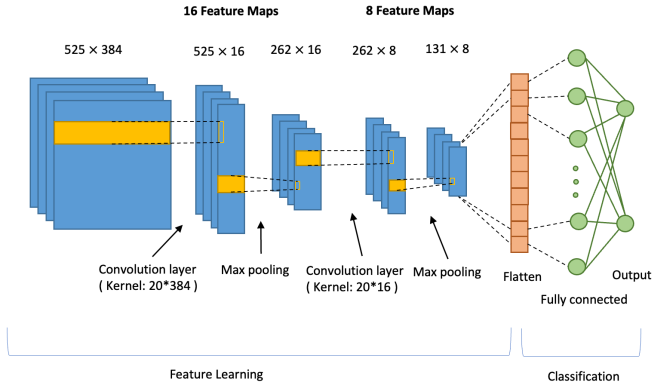

Fig. 2. Configuration of Convolutional Neural Network (CNN)

The general structure of CNN consists of convolutional layers, pooling layers, and fully connected layers. The convolution layer is in charge of discovering the most significant characteristics. A feature map is the result of the convolution operation, which is calculated by applying a feature detector (also called kernel or filter) to the input data. Pooling layer guarantees that the network can identify features regardless of where they are. Pooling such as max pooling or average pooling can also reduce the dimension of data given to CNN for computational efficiency and improve actual performance. The dropout layer is implemented to reduce overfitting. The pooled feature map is flattened into a single column by Flatten layer which is then linked to the fully connected layer. The result is transferred to an appropriate activation function before the final output is achieved.

Figure 2 shows the architecture of CNN used in the SBERT-CNN model. The input of CNN is a two-dimensional matrix with a shape of (N, 384) where N is the maximum number of posts per user. The CNN model consists of two 1D convolution components. Each of the components consists of a convolution layer, a max pooling layer, and a dropout layer. The kernel size of the first convolution layer is 20*384 and 20*16 for the second convolution layer. 16 and 8 convolutional filters were used in the first and second convolutional layers, respectively. The activation function ReLu is used in each convolutional layer for non-linear transformation. The higher-level characteristics can be extracted from the input matrix by multiple convolutional layers. The Max pooling layer is used to calculate the largest value from the feature map covered by the filter and reduce the feature map dimensionality by half. The dropout rates in the two convolutional components are all set to 20%. The flatten layer is fully connected to a hidden layer containing 32 neurons. The output layer has 2 neurons for 2 classes of output that is using the Softmax function. The optimizer function is Adam with a 1e-4 learning rate. The 'Sparse_categorical_crossentropy' is set as the loss function for parameter optimization. The model is trained and validated with 50 epochs.

### F. Experiments and Evaluation

The CNN model requires a consistent size of input matrix (N, 384) for each user but the number of posts per user varies among users, ranging from 40 to 80,498 in the depression dataset derived from SMHD. We set up a threshold for the maximum number of posts per user (N) in the input data of CNN to keep the size of input matrix of CNN to be consistent for all users. If the number of posts of a user was greater than the threshold N, we randomly selected the same number of posts of the user as the threshold. For the rest of the users with fewer posts than the threshold, we added vectors of zero to post embeddings. In order to evaluate the effectiveness of different number thresholds for posts per user in the input data of CNN, we tested the 50th, 75th, 80th, 90th, and 95th percentiles in the distribution of the number of posts for every user in depression and control groups for comparison. The experiment results of different thresholds of post count per user are listed in Table III.

We also experimented with the different number of epochs for parameter optimization and model performance. We calculated the loss function of training set and validation set in terms of optimization iterations. The results for 95th percentile are plotted in Figure 3.

To evaluate the performance of the proposed model, we used Accuracy, Precision, Recall, and F1 score as evaluation metrics as below:

*Accuracy*
$$= \frac{True\ Negative\ +\ True\ Positive}{True\ Negative\ +\ True\ Positive\ +\ False\ Negative\ +\ False\ Positive}$$

$$Precision = \frac{True\ Positive}{True\ Positive\ +\ False\ Positive}$$

$$Recall = \frac{True\ Positive}{True\ Positive\ +\ False\ Negative}$$

$$F1\ score = \frac{2\ *\ (Precision\ *\ Recall)}{Precision\ +\ Recall}$$

All experiments were performed on the Google Colabotorary (Colab for short) from Google Research, which allows the writing and executing of python codes in a browser. We implemented the CNN component in our hybrid SBERT-CNN model by using TensorFlow [40], a well-known, open-source software library for deep learning from Google. SentenceTransformers package with pre-trained model 'all-MiniLM-L6-v2' was used in our SBERT-CNN model as well [28].

### IV. RESULTS

#### A. Model Performance

Table II presents the model's performance of SBERT-CNN architecture compared to other machine learning and deep learning models. The traditional machine learning models (e.g., XGBoost, Logistic Regression, and linear SVM) with bag-of-word features and supervised FastTest showed poor performance of a binary classification for depression detection. The F1 score of those machine learning models was 0.43-0.54, which is around the random guessing. All the deep learning models including CNN, LSTM, and Hierarchical Attention Network (HAN) showed good performance with an F1 score of on less than 0.68. The performance of CNN was slightly higher than LSTM in terms of F1 score (0.79 vs 0.77). The SBERT-CNN model we proposed outperforms all other models (CNN,

LSTM, HAM and traditional machine learning models) with an 0.86 F1 score, 0.85 precision, and 0.87 recall at all the performance metrics. Compared with the state-of-the-art baselines (CNN) reported in the literature [23], the SBERT-CNN model increased the performance to predict Reddit users with depression by 7%.

TABLE II. PERFORMANCE OF SBERT-CNN COMPARED TO OTHER MODELS REPORTED IN THE LITERATURE

| Model | Accuracy | F1 score | Precision | Recall |
|---|---|---|---|---|
| SBERT-CNN | **0.86** | **0.86** | 0.85 | **0.87** |
| CNN [23] | NA | 0.79 | 0.72 | **0.87** |
| LSTM [23] | NA | 0.77 | 0.74 | 0.79 |
| XLNET [a] [26] | NA | 0.70 | 0.74 | 0.67 |
| BERT [a] [26] | NA | 0.68 | 0.73 | 0.66 |
| RoBERT [a] [26] | NA | 0.68 | 0.71 | 0.60 |
| HAN [30] | NA | 0.68 | NA | NA |
| Supervised FastText [24] | NA | 0.54 | 0.67 | 0.45 |
| Linear SVM [24] | NA | 0.53 | 0.79 | 0.4 |
| Logistic Regression [24] | NA | 0.43 | **0.85** | 0.29 |
| XGBoost [24] | NA | 0.43 | 0.82 | 0.29 |

[a] The results of XLNET, BERT, and RoBERT are reported in the post level instead of the user level.

### B. Optimization Iteration Experiment

Figure 3 illustrates the loss function of the training set in blue line and validation set in orange line over 50 epochs or iterations. The loss function of the training set continued to decrease as the optimization iterations increased and the loss function of validation set drops to a point around 10 iterations before growing again. As experience beyond that point demonstrates the overfitting and the inflection point in validation loss may be the stopped point at training. We used the early stopping technique to reduce the overfitting issue. The training process was set to terminate after 10 epochs.

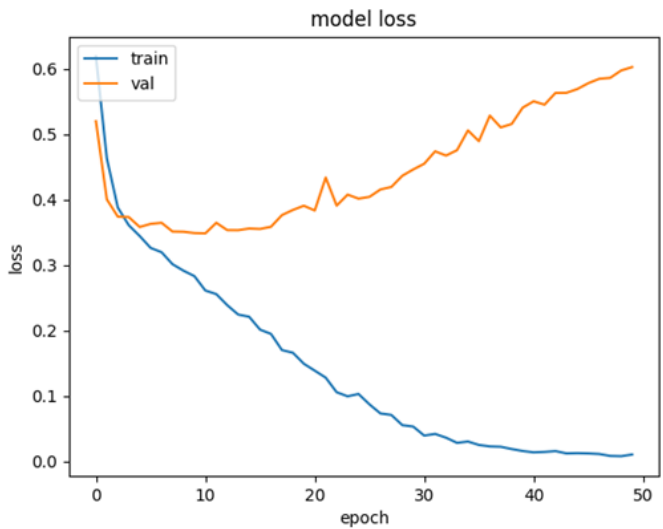

Fig. 3. Loss functions of the training set (blue) and validation set (orange) over epochs

### C. Post Number Experiment

Table III lists the model performance of SBERT-CNN model at different thresholds of posts count per user in the input matrix of CNN. The numbers in bold indicate the best performance for each performance measurement. The experiment results show the model performance (F1 score and accuracy) of SBERT-CNN model increased from 0.84 to 0.86 as the number threshold of posts per user rose from the 50th percentile to the 95th percentile. The performance for the 90th percentile and 95th percentile are very close. They both achieve an 0.86 F1 score and 0.86 accuracy. The precision at the 95th percentile is 0.01 higher but its recall is 0.02 lower than the 90th percentile. The results indicate that the model performance may converge at the 90th percentile and increasing the number of posts per user may not enhance the model performance.

TABLE III. PERFORMANCE OF SBERT-CNN IN TERMS OF THE NUMBER OF POSTS PER USER (N)

| Percentile | Posts/user | Accuracy | F1 score | Precision | Recall |
|---|---|---|---|---|---|
| 50th | 196 | 0.84 | 0.84 | **0.85** | 0.83 |
| 75th | 299 | 0.85 | 0.85 | **0.85** | 0.85 |
| 80th | 333 | 0.85 | 0.85 | 0.83 | 0.87 |
| 90th | 440 | **0.86** | **0.86** | 0.84 | **0.88** |
| 95th | 525 | **0.86** | **0.86** | **0.85** | 0.87 |

## V. DISCUSSION AND FUTURE WORK

Results of the comparative study in Table II showed that the hybrid model SBERT-CNN overcomes the state-of-art baseline performance produced by CNN in the literature using the depression datasets from the SHMD dataset [23]. Compared with CNN which took an input of word embedding matrix generated by Glove and word2vec techniques [41, 42], SBERT-CNN used the matrix of sentence embeddings generated by the pretrained SBERT from Reddit posts per user as input features. Word embedding is a low dimensional dense vector that encodes the meaning of the word. Words with similar meanings are close together in vector space. The word embedding is often used to create input features from textual data from machine learning models. Compared to word embedding, the sentence embedding generated from SBERT uses a low dimensional vector to encode complete sentences and their syntactic and semantic content. This can assist the machine learning models (e.g., CNN) in better comprehending the context and other details in the entire sentence. Thus, CNN based on SBERT showed a better performance than CNN based on Glove or word2vec.

With the same SMHD dataset, the results of transformer-based models (BERT [43], RoBERTa [44], and XLNET[46]) were reported in the post level rather than the user level with a F1-score of 0.68-0.70. Although the results of these transformers might not be comparable to the results in the user level, it would be worth to explore whether the performance of pure transformers (e.g., BERT) can be further improved by CNN via a hybrid neural network architecture in the future work. Jiang et al. constructed another dataset of Reddit posts for mental disorders, which is similar to SMHD dataset. They used the dataset to develop a binary classifier of depression identification with a F1 score of 0.84 in the user level, which is comparable to our result.

The results of optimization iteration experiment indicate overfitting. The overfitting of a model is characterized by a high performance on the training set but a low performance on the validation or testing set. The model has too much complexity and it captures some noise of the training data rather than the underlying trend. The model becomes too specific to the training data and loses its ability to generalize to new data. Overfitting can be prevented by using techniques such as regularization, early stopping [45], or cross-validation. In this work, we used early stopping technique to prevent overfitting and we will refine the structure of CNN and explore other techniques to prevent overfitting in the future.

We used a mini version of sentence BERT models ('all-MiniLM-L6-v2') in the SBERT-CNN to identify Reddit users with depression. We will explore other larger models of sentence BERT, particular, 'all-mpnet-base-v2', which has been widely tested as having the highest overall quality [28, 47]. We will also investigate different pretrained models such as domain specific language models to concatenate with CNN to further improve the performance of depression identification.

## VI. Conclusions

Depression is a common but serious mental disorder that affects hundreds of millions of people worldwide and is ranked as the leading cause of disability in the United State. It is critical to use prevention and intervention strategies including screening and early detection to reduce the influence of depression. Recently people with mental health issues are willing to use social media platforms (e.g., Reddit) to express their mental problems (e.g., depression) and seek support from other users, making social media a valuable resource for identifying individuals at risk. Many research works have been done for detecting depression from Reddit posts with NLP algorithms, machine learning methods, and deep learning techniques. The performance of the depression identifier is increasing as the development of machine learning methods, particularly, deep learning methods. In this work, we developed a hybrid neural network that integrates sentence BERT and CNN for identifying Reddit users with depression. The sentence BERT allows the learning of meaningful representation of each post; CNN enables the further transformation of those embeddings via convolution operation for identification of depression patterns of users. Comparison research utilizing the SMHD dataset revealed that the SBERT-CNN model outperforms the state-of-the-art baseline performance achieved by other models in the literature, suggesting the feasibility of the hybrid strategy. The hybrid model can apply to other text classification problems for other health problems.

## Corresponding Authors

Ming Huang (huang.ming@mayo.edu) is the corresponding author.

## Conflicts of Interest

The authors declare no conflict of interest.

## Acknowledgment

This research work was supported by the National Center for Advancing Translational Sciences (NCATS) at National Institutes of Health (NIH) (Grant Number UL1 TR002377). The content is solely the responsibility of the authors and does not necessarily represent the official views of the National Institutes of Health.